\documentclass[10pt,twocolumn,letterpaper]{article}

\usepackage{multirow}
\usepackage{algorithm}
\usepackage{algorithmic}
\usepackage{graphicx}
\usepackage[table,xcdraw]{xcolor}
\usepackage{booktabs}   
\usepackage{amsmath} 
\usepackage{amsmath}
\usepackage{amssymb}

\definecolor{wacvblue}{rgb}{0.21,0.49,0.74}
\usepackage[pagebackref,breaklinks,colorlinks,allcolors=wacvblue]{hyperref}

\date{} 
\title{Model-Agnostic Gender Bias Control for Text-to-Image Generation via Sparse Autoencoder\thanks{Under Review}}

\author{
Chao Wu$^1$, Zhenyi Wang$^2$, Kangxian Xie$^1$, Naresh Kumar Devulapally$^1$,\\ 
Vishnu Suresh Lokhande$^1$, Mingchen Gao$^{1}$\thanks{Corresponding author.} \\
$^1$University at Buffalo, USA \\
$^2$University of Maryland, College Park, USA \\
\texttt{\{cwu64, kangxian, devulapa, vishnulo, mgao8\}@buffalo.edu},\\
\texttt{zwang169@umd.edu}
}

\begin{document}

\maketitle
\begin{abstract}
Text-to-image (T2I) diffusion models often exhibit gender bias, particularly by generating stereotypical associations between professions and gendered subjects. This paper presents SAE Debias, a lightweight and model-agnostic framework for mitigating such bias in T2I generation. Unlike prior approaches that rely on CLIP-based filtering or prompt engineering, which often require model-specific adjustments and offer limited control, SAE Debias operates directly within the feature space without retraining or architectural modifications. By leveraging a k-sparse autoencoder pre-trained on a gender bias dataset, the method identifies gender-relevant directions within the sparse latent space, capturing professional stereotypes. Specifically, a biased direction per profession is constructed from sparse latents and suppressed during inference to steer generations toward more gender-balanced outputs. Trained only once, the sparse autoencoder provides a reusable debiasing direction, offering effective control and interpretable insight into biased subspaces. Extensive evaluations across multiple T2I models, including Stable Diffusion 1.4, 1.5, 2.1, and SDXL, demonstrate that SAE Debias substantially reduces gender bias while preserving generation quality. To the best of our knowledge, this is the first work to apply sparse autoencoders for identifying and intervening in gender bias within T2I models. These findings contribute toward building socially responsible generative AI, providing an interpretable and model-agnostic tool to support fairness in text-to-image generation. 
\end{abstract}
    
\section{Introduction}
\label{sec:intro}

Text-to-image (T2I) diffusion models \cite{rombach2022high,podell2023sdxl,ramesh2022hierarchical,mei2024codi} have shown significant progress in generating images with high fidelity and resolution in recent studies. Beyond general text-to-image generation, recent work has also explored
fine-grained controllable or instruction-driven editing in both diffusion and
autoregressive models \cite{lan2025flux,xu2025scalar,jin2025scar},
demonstrating the broader trend of improving semantic controllability at
inference time.
However, potential bias may exist in the generating process regarding sensitive attributes such as gender, race, and age \cite{luccioni2023stable,anivcin2022bias,wu2023stable}. Among these, gender bias has long been present and remains challenging to eliminate, as it is often implicit and embedded in various semantic contexts. Due to the different social roles traditionally assigned to men and women, gender bias becomes particularly evident in workplace scenarios, and such patterns are learned and reinforced by T2I models \cite{seshadri2023bias}. In particular, professional identity prompts frequently expose gender biases in T2I models \cite{wu2024stable,girrbach2025large}, which reflect and reinforce real-world stereotypes. Controlling such biases during generation remains a challenging problem, particularly without the need for costly model retraining or fine-tuning.
\begin{figure}[t!]
    \centering

    \includegraphics[width=\linewidth]{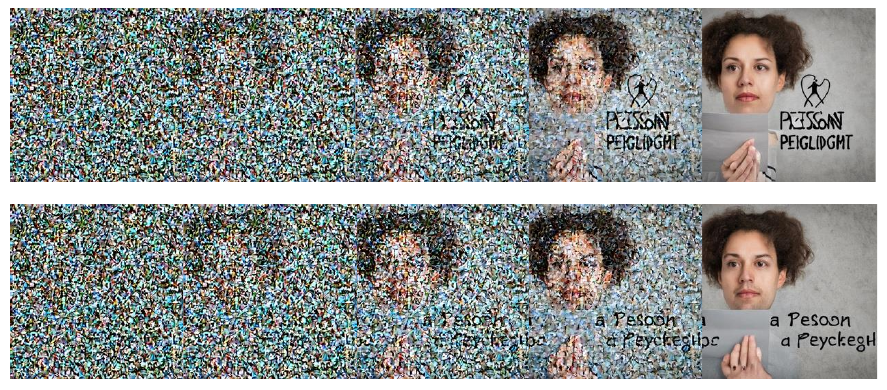}
    \caption{Comparison of Before (Top) and After (Bottom) SAE Debiasing of “a photo of a person who works as a psychologist”.}
    \label{fig:psychologist}
\end{figure}

Prior efforts to mitigate gender bias in text-to-image generation can be broadly categorized into training-based and training-free approaches. Training-based methods involve retraining or fine-tuning the model or adding trainable components, which are costly and usually not model-agnostic \cite{gandikota2024unified,kim2023stereotyping,orgad2023editing,seth2023dear}. Other training-free methods \cite{wang2021gender,chuang2023debiasing,friedrich2023fair,jung2024unified} focus on modifying prompts or simply removing gender-related features from the text encoder, but these often lack fine-grained control and may result in the loss of semantic fidelity. Moreover, they tend to overlook the interdependence between features and the inherent ambiguity of semantics, which is particularly problematic when addressing subtle and implicit forms of bias, such as gender bias. 

To address these limitations, we propose \textit{SAE Debias}, a lightweight, model-agnostic framework for mitigating gender bias in profession-related prompts (Figure~\ref{fig:psychologist}). Our method requires only a one-time training of a sparse autoencoder (SAE) \cite{gao2024scaling} on hidden states from a frozen CLIP-based text encoder. During inference, we identify and subtract gender bias directions in the sparse space—without modifying the original model—thus achieving targeted debiasing with minimal impact on image quality.
Our contributions can be summarized as follows:

\begin{itemize}
  \item \textbf{Training efficiency.} Our framework requires training only once on a lightweight SAE. It operates entirely at inference time and can be easily integrated into existing text-to-image pipelines without additional fine-tuning.

  \item \textbf{Model-agnostic design.} Our method is model-agnostic and can be applied to diffusion models that utilize CLIP-family text encoders, without requiring any model-specific adjustments.

  \item \textbf{Residual-based intervention.} Instead of trimming or interpolating features, our method performs residual adjustments on the hidden states of the text encoder during inference, preserving more of the original semantic information while maintaining the consistency of the generated images.
\end{itemize}

\section{Related Works}
\label{sec:related}
\subsection{Gender Bias in Stable Diffusion and Debiasing Approaches}
Text-to-Image Diffusion Models are inherently biased as they are trained on large datasets that are sampled in real-world settings \cite{schuhmann2022laion}. Since many T2I models, such as Stable Diffusion (SD) and DALL·E 2 \cite{ramesh2022hierarchical}, rely on CLIP (Contrastive Language-Image Pre-Training) \cite{radford2021learning} for text-image alignment, biases in CLIP can propagate into the generated images \cite{wolfe2022evidence}. \cite{radford2021learning} found CLIP model exhibits biases against certain protected attributes, and \cite{wolfe2022evidence} observed that CLIP directs more attention to the physical features of women (such as hair) and exhibits substantial racial and gender bias. Biases in both the text encoder and the training data can propagate into diffusion models, leading to stereotypical outputs during the generation process \cite{buolamwini2018gender}. As a result, recent research has increasingly focused on identifying and mitigating gender bias in diffusion models.

A common strategy is model retraining or fine-tuning. Esposito et al. \cite{esposito2023mitigating} fine-tune Stable Diffusion models using debiased synthetic datasets. \cite{teo2023fair} leverages transfer learning to adapt biased generative models toward fair distributions by fine-tuning on small, unbiased datasets. DeAR \cite{seth2023dear} extends adversarial training by introducing an additive residual branch to neutralize sensitive attributes. However, such retraining-based approaches are often computationally expensive and lack adaptability to new tasks.

Other methods aim to steer internal representations without retraining. SFID \cite{jung2024unified} applies low-confidence feature interpolation to trim gender-related features. CLIP-Clip \cite{wang2021gender} prunes latent features with high mutual information with protected attributes. \cite{chuang2023debiasing} relies on predefined prompts to mitigate gender bias using a calibrated projection matrix. These methods offer more lightweight solutions, but operations such as direct interpolation and pruning can be overly aggressive, disturbing feature representations and degrading generation quality.
\subsection{Sparse Autoencoders for Interpretability and Bias Control}
\label{sae}
Sparse Autoencoders (SAEs) \cite{templeton2024scaling,wang2024interpret} have recently emerged as a powerful tool for enhancing interpretability in large models, particularly language and vision-language models. In LLM, SAEs address the inherent polysemanticity problem in neural representations, where a single neuron responds to multiple unrelated concepts \cite{huben2023sparse}. This phenomenon has also been extended to the domain of VLM in \cite{pach2025sparse} recently to fetch visual concepts by selecting the highest activating images in sparse latents of SAEs. 

Meanwhile, SAEs offer practical advantages for feature space interventions without retraining the full model. They enable fine-grained feature space level interventions without retraining the whole model. Modifying activations of specific neurons can steer model outputs toward desired attributes or suppress unwanted ones \cite{kim2025concept}. This makes SAEs particularly suited for debiasing tasks, where bias-inducing components (e.g., gender, race) can be identified and controlled without impacting unrelated features.
 
\section{Preliminaries}
\subsection{Sparse Autoencoder}
\label{detail}
Sparse Autoencoders \cite{ng2011sparse} are unsupervised neural networks consisting of an encoder and a decoder that are optimized jointly using reconstruction error and a sparsity regularization term. The sparsity constraint on the latent activations encourages the network to activate only a small subset of neurons for any given input. These sparsely activated neurons are referred to as sparse latent variables. The sparsity constraint encourages only a few neurons to be activated for a given input so that the most important representations are captured during optimization. The k-sparse autoencoder \cite{makhzani2013k} directly controls sparsity by only keeping the k largest latents and zeroing the rest, providing a fast way to converge in large models. 

\label{subsec:train_sae}
Given the hidden states extracted from the text encoder of the SD model, k-SAE is trained to obtain a sparse latent representation, which captures patterns related to gender and profession in our case. Given the residual features $z \in \mathbb{R}^{n \times d}$ extracted from the $(L-1)$-th layer of the text encoder, the k-SAE encoder $f(\cdot)$ maps $z$ into a sparse latent code $h$:
\begin{equation}
h = f(z) = \text{Top-}k(\text{ReLU}((W_{\text{enc}} z -  b_{\text {pre}}) + b_{\text{enc}}),
\end{equation}
where $W_{\text{enc}} \in \mathbb{R}^{m \times d}$, $b_{\text{enc}} \in \mathbb{R}^{m}$ and are the encoder weights and bias, $b_{\text {pre}} \in \mathbb{R}^{d}$ is actually the decoder bias which is initialized as the geometric median of the dataset according to 
\cite{bricken2023monosemanticity} to facilitate the recovery of sparse features.
$m/d$ is referred to as the \textit{expansion factor} controlling the hidden layer size relative to the input dimension, and $\text{Top-}k(\cdot)$ retains only the $k$ largest activations \cite{makhzani2013k} and sets the rest to zero.

The decoder $g(\cdot)$ reconstructs $z$ from $h$ via a linear projection:
\begin{equation}
\hat{z} = g(h) = W_{\text{dec}} h + b_{\text{pre}},
\end{equation}
where $W_{\text{dec}} \in \mathbb{R}^{d \times m}$ is the decoder weight.
The primary training objective is to minimize the mean squared error (MSE) reconstruction loss between the input $z$ and reconstructed $\hat{z}$, along with an auxiliary loss \cite{gao2024scaling}(see supplementary material) that encourages underutilized latents to contribute.
\begin{equation}
\mathcal{L}_{\text{MSE}} = \frac{1}{n} \sum_{i=1}^{n} \| z_i - \hat{z}_i \|_2^2.
\end{equation}
\begin{equation}
\mathcal{L}_{\text{SAE}} = \mathcal{L}_{\text{MSE}} + \alpha \mathcal{L}_{\text{aux}},
\end{equation}

\subsection{Bias in Bios Dataset}
We train our k-SAE on the Bias in Bios dataset~\cite{de2019bias}, a widely-used benchmark for evaluating gender bias in natural language processing models. The training split of Bias in Bios consists of 257,478 short biographies, each labeled with one of 28 professional occupations (e.g., physician, nurse, software engineer) and a binary gender attribute inferred from pronouns. The biographies were collected from Common Crawl~\cite{commoncrawl}, an open-access repository of large-scale web-scraped data. The dataset is specifically designed to study how occupational classification models exhibit gender biases.

For our purposes, we use the hard-text version of the biographies from the training split, where gender-indicative terms such as names and pronouns are preserved. We extract the hard texts as prompts, which are then fed into the text encoder of the Stable Diffusion model. Our goal is to extract the gender-relevant direction for each profession based on the residual features associated with the corresponding text embeddings. As observed in the supplementary material, the distribution of Bias in Bios dataset is highly imbalanced, with several professions dominated by a specific gender. Motivated by this observation, we extract gender bias directions conditioned on each profession in our method to enable bias control.

\section{SAE Debias Framework}
\begin{figure*}[!htbp]
    \centering
    \includegraphics[width=0.95\linewidth]{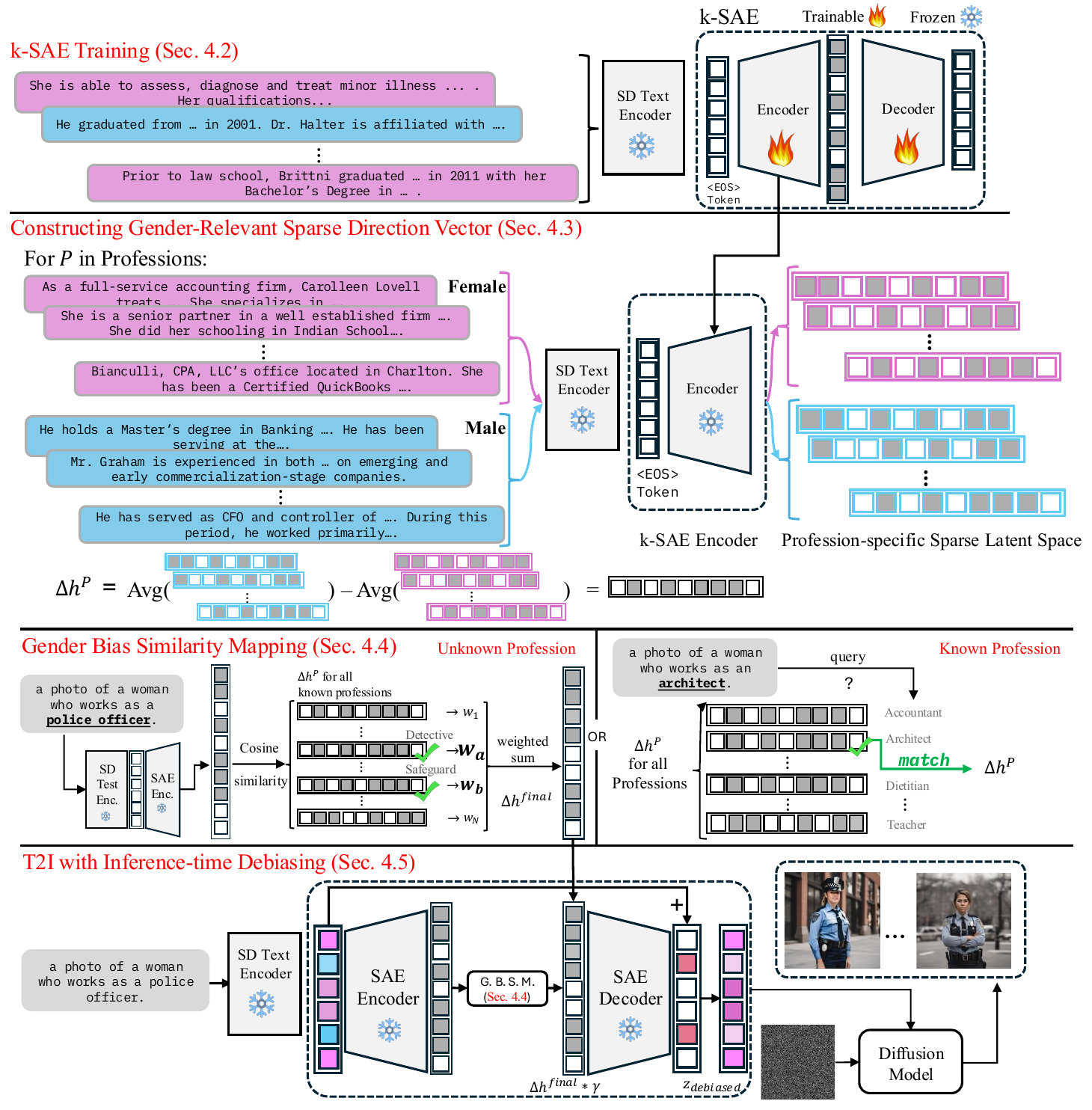} % 缩到 90%
    \caption{Pipeline overview for SAE Debias.}
    \label{fig:pipeline}
\end{figure*}

In this section, we cover the details of our proposed debias pipeline. Building on the intermediate hidden states extracted from the penultimate layer of the CLIP-based text encoder(Sec. ~\ref{feature}), we first train a k-SAE network to learn sparse latent representations for profession and gender-related texts~(Sec.~\ref{train_sae}). With the pretrained k-SAE, we locate the gender-bias direction in the sparse latent space~(Sec.~\ref{subsec:gender_dir}), and finally perform debiasing~(Sec.~\ref{mapping} and Sec.~\ref{subsec:inference_debias}) by steering the text embedding in the gender-neutral direction for SD generation.
\subsection{Feature Extraction}
\label{feature}
We extract the penultimate layer output from the text encoder of a pre-trained Stable Diffusion model. Specifically, we hook the second-to-last transformer layer, whose output is rich in semantics and often used as input to the cross-attention mechanism during image generation.
\begin{equation}
H = f_{\text{text}}(X),
\label{eq:text_encoding}
\end{equation}
where $H = \{ h_1, h_2, \ldots, h_n \} \in \mathbb{R}^{n \times d}$, with $n$ tokens and hidden dimension $d$.
We extract the residual features $z$ from the output of the $(L-1)$-th layer:
The input prompts $X$ are from the Bias in Bios dataset~\cite{de2019bias}, which consists of textual biographies labeled with binary gender attributes and professional occupations. Specifically, we use the \textit{hard text} column of each biography as the prompt.
\begin{equation}
z = H^{(L-1)},
\label{eq:residual_features}
\end{equation}
where $H^{(L-1)}$ denotes the hidden states at the second-to-last layer. $z$ is the residual feature from the second-to-last layer's hidden state at the position corresponding to the [EOS] token, which summarizes the entire prompt.
\subsection{Training k-sparse autoencoders}
\label{train_sae}
We train the k-SAE described in Section~\ref{subsec:train_sae} to learn sparse latent representations from the SD text encoder. Specifically, given a text prompt, we extract the hidden state corresponding to the end-of-sentence (EOS) token from the penultimate transformer layer of the SD text encoder. These hidden features $z \in \mathbb{R}^d$ serve as the inputs to the SAE.

During training, we enforce top-$k$ sparsity on the encoder outputs by retaining only the $k$ largest activations. We experiment with different $k$ values (\(k \in \{32, 48, 64\}\)) for different SDs to control the level of sparsity with regard to model size. Specifically, we set the sparsity level to $k=32$ and the expansion factor to $32\times$ for Stable Diffusion 1.x models. For 2.x models, the sparsity level remains $k=32$, but the expansion factor is increased to $48\times$. For SDXL, we adopt $k=64$ with an expansion factor of $64\times$ to accommodate its larger model architecture.

The k-SAE is trained using the combined reconstruction and auxiliary losses defined in Section 3.1. We set the auxiliary loss weight $\alpha$ to $1/32$ as suggested by \cite{kim2025concept} throughout all experiments.
During inference, we remove the top-$k$ operation, allowing continuous steering of the latent activations without hard sparsity constraints \cite{kim2025concept}, as the Top-K function disregards certain minor activations that may contain important information for both maintaining visual quality and effective debiasing. 

\subsection{Gender-Relevant Sparse Direction}
\label{subsec:gender_dir}
This section explains the operation for constructing the gender-relevant sparse direction for all professions in the training set. 
Given prompts from the Bias in Bios dataset, following the pipeline in Sec.~\ref{subsec:train_sae}, we acquire the sparse latent feature for each prompt using our trained k-SAE. As each prompt is associated with a gender label (male or female) and a profession label \(P\), for each profession, we could compute the average sparse latent for male and female:
\begin{equation}
\mu_{\text{male}}^{(P)} = \frac{1}{N_m} \sum_{i=1}^{N_m} f\left( z_{\text{male},i}^{(P)} \right),
\end{equation}
\begin{equation}
\mu_{\text{female}}^{(P)} = \frac{1}{N_f} \sum_{j=1}^{N_f} f\left( z_{\text{female},j}^{(P)} \right),
\end{equation}
where \( N_m \) and \( N_f \) denote the number of male and female samples for profession \( P \), respectively.

The gender-relevant direction for profession \( P \) is then defined as the difference between the male and female means:

\begin{equation}
\Delta h^{(P)} = \mu_{\text{male}}^{(P)} - \mu_{\text{female}}^{(P)}.
\end{equation}

This direction captures the stereotypical shift between genders conditioned on the profession.

\subsection{Gender Bias Similarity Mapping}
\label{mapping}
At inference time, given a new prompt, we project the residual features of the job token into the sparse latent space using the k-SAE we trained and compute their similarity to precomputed gender-relevant sparse directions to mitigate bias. The similarity computation focuses on matching the subtle bias-related components with the prompts rather than purely isolating professional information, allowing for effective debiasing while preserving the original profession semantics. Our approach handles two cases:

\textbf{Known professions:} If the profession in the prompt matches one of the professions we calculated during Sec.~\ref{subsec:gender_dir}, we directly retrieve the corresponding gender-relevant sparse direction \( \Delta h^{(P)} \) as computed in Sec.~\ref{subsec:gender_dir} as $\Delta h^{\text{final}}$.

\begin{equation}
\Delta h^{\text{final}} = \Delta h^{(P)}.
\end{equation}

\textbf{Unseen professions:} If the profession is not in the pre-computed list, we infer a debiasing direction dynamically. Specifically, we extract the residual feature \( z_{\text{job}} \) corresponding to the job token and get its sparse latent  \( h_{\text{job}} \).Then we compute its cosine similarity with each pre-computed direction \( \Delta h^{(P_i)} \). The similarities are normalized using a temperature-scaled softmax:

\begin{equation}
w_i = \frac{\exp\left(\text{Sim}(h_{\text{job}}, \Delta h^{(P_i)}) / T\right)}{\sum_j \exp\left(\text{Sim}(h_{\text{job}}, \Delta h^{(P_j)}) / T\right)},
\end{equation}
Here, \( i \) refers to each stored profession individually when computing its similarity and weight, while \( j \) sums over all stored professions to normalize the softmax. Specifically, \( i \) selects the profession whose weight \( w_i \) is being calculated, and \( j \) ensures that the softmax denominator covers all professions for proper normalization.
where \( T \) is a temperature hyperparameter, and \( \text{Sim}(\cdot, \cdot) \) denotes cosine similarity.

The final steering direction is computed as a weighted sum: 

\begin{equation}
\Delta h^{\text{final}} = \sum_i w_i \Delta h^{(P_i)}.
\end{equation}
\subsection{Inference-time Debiasing}
\label{subsec:inference_debias}
Finally, we steer the residual feature \( z \) by applying a small shift along \( \Delta h^{\text{final}} \) using the trained SAE decoder:

\begin{equation}
z_{\text{debiased}} = z + W_{\text{dec}}(\gamma \cdot \Delta h^{\text{final}}),
\end{equation}

where \( \gamma \) is a small negative scaling coefficient controlling the strength of the debiasing. During SAE training, residual features were extracted at the [EOS] token to capture sentence-level gender bias implicitly associated with professions. During inference, we apply the intervention at the residual corresponding to the job token to achieve localized bias mitigation without disrupting the overall semantics.

\section{Experiments}
\subsection{Setup}
Unlike previous studies on Stable Diffusion debiasing\cite{jiang2024mitigating} \cite{he2024debiasing} \cite{huang2025debiasing} that focus on a limited set of model instances, we evaluate our method across multiple versions, including Stable Diffusion 1.4, 1.5, 2.1, and SDXL\footnote{We use \texttt{stabilityai/stable-diffusion-xl-base-1.0} for SDXL.}. For each model, we extract residual features from the text encoder by hooking the output of the second-to-last transformer layer ($L-1$ layer)\footnote{For SDXL, we hook the only main text encoder with 1280 dimensions and the auxiliary text encoder remains unchanged}. These residual activations serve as the input to train a $k$-sparse autoencoder.

To systematically control and evaluate for gender bias, we utilize the input prompt from  \cite{cho2023dall} to generate images. we construct text prompts following the template: ``a photo of $\mathcal{G}$ who works as a[n] $\mathcal{P}$,'' where the gender set $\mathcal{G} = \{\text{a man}, \text{a woman}, \text{a person}\}$ . We use 83 professions as in \cite{cho2023dall}

\subsection{Evaluation Metrics}

As discussed in \cite{jung2024unified}, bias in text-to-image models manifests in two major ways: generating the wrong gender for a prompt with a specific gender and generating biased images for a gender-neutral prompt. We evaluate our method under both of these settings. 
First, we adopt the approach in \cite{jung2024unified} to detect the gender of generated images using BLIP-2 \cite{li2023blip} by asking the binary question: \emph{``Does the person look like a male or a female?''}
and BLIP-2 returns the detected answer.
For prompts that explicitly specify a gender, a fair text-to-image generation model should generate images matching the intended gender regardless of the profession. To quantify this, we compute the \textbf{Mismatch Rate} as:

\begin{equation}
\frac{1}{|\mathcal{D}|} \sum_{k \in \mathcal{D}} I_k,
\end{equation}

where $\mathcal{D}$ is the set of prompts and $I_k$ is an indicator variable that equals 1 if the generated image has a gender prediction different from the gender in the prompts.

To compare results with \cite{jung2024unified}, we also calculate the \textbf{Composite Misclassification Rate} $M_{RC}$ by averaging the mismatch rates across male and female subgroups, ensuring a balanced evaluation between genders.
\begin{equation}
MR_C = \sqrt{MR_O^2 + (MR_F - MR_M)^2}
\end{equation}

where $MR_O$ denotes the overall mismatch rate across all samples, and $MR_F$ and $MR_M$ represent the mismatch rates for female and male subgroups, respectively.

The composite mismatch rate, which combines the overall misclassification captured by $MR_O$ and the disparity between genders captured by the absolute difference $(MR_F - MR_M)$, penalizes models that achieve low overall error rates while exhibiting significant discrepancies between male and female groups.

Second, to assess debiasing in gender-neutral prompts, we evaluate the model’s fairness using prompts like we constructed ``a photo of a person who works as a[n] $\mathcal{P}$ ''. Fairness requires the model to generate a relevantly balanced number of male and female images across all professions.  We measure the skewness in image distribution using the \textbf{Skew} metric, defined as:

\begin{equation}
\text{Skew} = \frac{1}{|\mathcal{P}|} \sum_{p \in \mathcal{P}} \frac{\max(N_{p,m}, N_{p,f})}{C},
\end{equation}

where $\mathcal{P}$ is the set of professions, $N_{p,m}$ and $N_{p,f}$ are the number of generated male and female images for profession $p$, and $C$ is the total number of generations per profession (we set $C=10$ in our experiments for precise comparison).

Lower values of mismatch rates and \textit{Skew} indicate better fairness performance. We report results across multiple Stable Diffusion models, comparing our method with previous baselines and recent debiasing approaches.

\subsection{Results}
We present the experimental results of our method on different Stable Diffusion models. We report mismatch rates and skewness as percentages (\%), where lower values indicate better fairness performance and evaluate our method on multiple versions of Stable Diffusion models (SD 1.4, SD 1.5, SD 2.1, and SDXL). 
Table ~\ref{tab:main_results} reports the mismatch rates for gender-specific prompts and neutral prompt skewness, comparing our approach with the baseline models. On SD 1.4, our method significantly reduces the mismatch rate across all metrics, achieving 0.06 overall mismatch compared to the baseline 0.84. Similarly, on SD 1.5, although the baseline already achieves a relatively low male mismatch rate, our method further improves the female mismatch and overall fairness. For SD 2.1, we observe that our method consistently lowers the overall mismatch rate from 0.78 to 0.60 and improves the neutral prompt skewness.

On SDXL, since the baseline already achieves a perfect mismatch rate on gender prompts, our method does not improve mismatch rates but reduces the skewness by over 6\%. Interestingly, our method is especially effective on earlier versions (SD 1.4, SD 1.5), where gender bias may be more pronounced. 

As shown in Table~\ref{comparison}, our method consistently outperforms other approaches across all metrics on SD 1.4 and achieves the best skew values across all models. It obtains either the lowest or the second-lowest mismatch rates for both male and female prompts, reaching an overall mismatch rate of just $0.06\%$ on SD 1.4 while maintaining strong performance on SD 1.5 and SD 2.1.

Compared to SFID\cite{jung2024unified} and CLIP-clip\cite{wang2021gender}, our method significantly reduces gender bias without compromising overall performance. Notably, Prompt Debias\cite{chuang2023debiasing} not only exhibits high mismatch rates on earlier versions but also severely degrades mismatch performance across all categories in SD 2.1 due to overcompensation toward skewness.

Furthermore, our method achieves consistently lower skew values on neutral prompts, indicating improved fairness in gender-neutral generation. These results demonstrate that our approach provides a stable and effective debiasing solution across different Stable Diffusion versions.

\begin{table*}[t]
\caption{Experimental results for text-to-image generation. \textbf{Bold} indicates the best result for each baseline.}
    \vspace{1em}  % 你可以调成 0.5em、0.7em 试试效果
\label{tab:main_results}
\centering
\small
\setlength{\tabcolsep}{1.8pt}
\renewcommand{\arraystretch}{1.1}
\begin{tabular}{llccccc}
\toprule
    \textbf{Model} & & \multicolumn{4}{c}{\textbf{Mismatch Rate~$\downarrow$}} & 
     \\
    & & Male & Female & Overall & Composite & \textit{Skew~$\downarrow$} \\
\midrule
\multirow{2}{*}{SD 1.4} 
& Base & 1.45$\pm$0.95 & 0.24$\pm$0.51 & 0.84$\pm$0.42 & 1.74$\pm$0.92 & 85.18 \\
& Ours & \textbf{0.12$\pm$0.39} & \textbf{0.00$\pm$0.00} & \textbf{0.06$\pm$0.19} & \textbf{0.14$\pm$0.43} & \textbf{83.49} \\
\midrule
\multirow{2}{*}{SD 1.5} 
& Base & \textbf{0.60$\pm$0.63} & 0.24$\pm$0.51 & 0.42$\pm$0.29 & 0.94$\pm$0.65 & 83.25 \\
& Ours & 0.60$\pm$0.85 & \textbf{0.12$\pm$0.38} & \textbf{0.36$\pm$0.51} & \textbf{0.66$\pm$0.94} & \textbf{81.93} \\
\midrule
\multirow{2}{*}{SD 2.1} 
& Base & 1.45$\pm$1.24 & \textbf{0.12$\pm$0.38} & 0.78$\pm$0.64 & 1.60$\pm$1.39 & 83.01 \\
& Ours & \textbf{0.52$\pm$0.84} & 0.48$\pm$0.84 & \textbf{0.60$\pm$0.63} & \textbf{1.01$\pm$0.98} & \textbf{82.53} \\
\midrule
\multirow{2}{*}{SDXL} 
& Base & \textbf{0.00$\pm$0.00} & \textbf{0.00$\pm$0.00} & \textbf{0.00$\pm$0.00} & \textbf{0.00$\pm$0.00} & 93.98 \\
& Ours & 1.69$\pm$1.52 & 0.24$\pm$0.51 & 0.96$\pm$0.86 & 1.81$\pm$1.63 & \textbf{87.95} \\
\bottomrule
\end{tabular}
\end{table*}

\newcommand{\cmark}{\textcolor{green}{\ding{51}}}
\newcommand{\xmark}{\textcolor{red}{\ding{55}}}
\definecolor{darkred}{HTML}{EBFFEA}

\begin{table*}[!htbp]

  \caption{Comparison of Debiasing Methods. \textbf{Bold} indicates the best result for each model, while \underline{underline} denotes the second-best result.}
  \vspace{1em}  % 你可以调成 0.5em、0.7em 试试效果
  \label{tab:debiasing_results}
  \centering
  \small 
  \begin{tabular}{llccccc}
    \toprule
    \textbf{Model} & \textbf{Method} & \multicolumn{4}{c}{\textbf{Mismatch Rate (Gender Prompt)~$\downarrow$}} & \textbf{Neural Prompt} \\
    & & Male & Female & Overall & Composite & \textit{Skew~$\downarrow$} \\
    \midrule
    \multirow{4}{*}{SD 1.4} 
    & SFID       & 1.20$\pm$1.27 & \underline{0.12$\pm$0.38} & 0.66$\pm$0.60 & 1.48$\pm$1.34 & \underline{84.58} \\
    & Prompt Debias   & \underline{0.24$\pm$0.51} & 99.64$\pm$0.58 & 49.94$\pm$0.19 & 111.24$\pm$0.94 & 99.76 \\
    & CLIP-clip       & 1.08$\pm$1.33 & \textbf{0.00$\pm$0.00} & \underline{0.54$\pm$0.66} & \underline{1.21$\pm$1.48} & 84.70 \\
     & \cellcolor{red!20} Ours      & \cellcolor{red!20}\textbf{0.12$\pm$0.39} & \cellcolor{red!20}\textbf{0.00$\pm$0.00} & \cellcolor{red!20}\textbf{0.06$\pm$0.19} &  \cellcolor{red!20}\textbf{0.14$\pm$0.43} &  \cellcolor{red!20}\textbf{83.49}  \\
    \midrule
    \multirow{4}{*}{SD 1.5} 
    & SFID       & 1.41$\pm$0.49 & \textbf{0.00$\pm$0.00} & 0.70$\pm$0.25 & 1.57$\pm$0.55 & \underline{83.68} \\
    & Prompt Debias   & \textbf{0.52$\pm$0.95} & 99.6$\pm$0.85 & 43.84$\pm$15.05 & 110.57$\pm$2.03 & 99.70 \\
    & CLIP-clip       & 1.08$\pm$1.20 & \underline{0.12$\pm$0.38} & \underline{0.60$\pm$0.75} & \underline{1.15$\pm$1.20} & 83.37 \\
    & \cellcolor{red!20}Ours      & \cellcolor{red!20}\underline{0.60$\pm$0.85} & \cellcolor{red!20}\underline{0.12$\pm$0.38} & \cellcolor{red!20}\textbf{0.36$\pm$0.51} & \cellcolor{red!20}\textbf{0.66$\pm$0.94} & \cellcolor{red!20}\textbf{81.93}\\
    \midrule
    \multirow{4}{*}{SD 2.1} 
    & SFID       & {1.20$\pm$0.98} & \textbf{0.48$\pm$0.62} & 0.84$\pm$0.65 & 1.25$\pm$1.04 & \underline{83.86} \\
    & Prompt Debias   & 8.92$\pm$5.94 & 15.39$\pm$11.68 & 11.39$\pm$5.75 & 17.84$\pm$11.33 & 76.57 \\
    & CLIP-clip       & \underline{0.60$\pm$0.63} & 0.36$\pm$0.58 & \textbf{0.48$\pm$0.38} & \textbf{0.93$\pm$0.64} &84.70 \\

     & \cellcolor{red!20}Ours & \cellcolor{red!20}\textbf{0.52$\pm$0.84} & \cellcolor{red!20}\underline{0.48$\pm$0.84} & \cellcolor{red!20}\underline{0.60$\pm$0.63} & \cellcolor{red!20}\underline{1.01$\pm$0.98} & \cellcolor{red!20}\textbf{82.53} \\
    \bottomrule
  \end{tabular}
  \label{comparison}
\end{table*}

\subsection{Ablation Studies}
We first conduct an ablation study to investigate different strategies for computing the debiasing direction $\Delta h^{P}$. In the main approach, the debiasing direction is computed based on the sparse features extracted from the job token states. We compare it against two variants:

\textbf{EOS-based Direction}: Instead of using the job token, we use the sparse latent corresponding to the end-of-sentence (EOS) token states, which encodes the entire prompt. The same cosine similarity and softmax weighting are applied.
    
\textbf{Profession-average Direction}: In getting the profession-wise gender-bias direction, we compute $\Delta h^{P}$ by taking the average of all male and female latent features per profession, instead of the difference between male and female.

As shown in Table~\ref{ablation}, using the EOS states for bias direction computation leads to higher mismatch rates and skew values, indicating that relying on the global prompt representation weakens the ability to capture localized bias information. Moreover, even computing the bias direction by averaging profession-specific sparse features achieves comparable or slightly better skew values on SD 1.5 and SD 2.1, it consistently results in higher mismatch rates across all SD models. We hypothesize that averaging male and female latent representations overly smooths the gender-specific features, which reduces skewness but makes the generated samples less distinguishable in terms of gender, leading to an increase in mismatch rates. This is undesirable, as we aim to mitigate gender bias without reducing specific gender attributes in the generated outputs.

\begin{table*}

  \centering
  \caption{Comparison of Different SAE Debiasing Strategies on SD Models.}
    \vspace{1em}  % 你可以调成 0.5em、0.7em 试试效果
  \label{tab:sae-debias}
  \vspace{-0.6em}
  \small
  \setlength{\tabcolsep}{2pt}
  \renewcommand{\arraystretch}{1.1}
  \begin{tabular}{llccccc}
    \toprule
     & \textbf{Method} & \multicolumn{4}{c}{\textbf{Mismatch Rate}~$\downarrow$} &\\
    & & Male & Female & Overall & Composite & \textit{Skew~$\downarrow$} \\
    \midrule
    \multirow{3}{*}{1.4 } & Ours & 0.12$\pm$0.39 & 0.00$\pm$0.00 & 0.06$\pm$0.19 & 0.14$\pm$0.43 & 83.49 \\
    & EOS & 1.08$\pm$0.89 & 0.24$\pm$0.76 & 0.66$\pm$0.60 & 1.29$\pm$1.04 & 83.73 \\
    & Avg. & 0.72$\pm$0.84 & 0.00$\pm$0.00 & 0.36$\pm$0.42 & 0.81$\pm$0.94 & 84.94 \\
    \midrule
    \multirow{3}{*}{1.5 } & Ours & 0.60$\pm$0.85 & 0.12$\pm$0.38 & 0.36$\pm$0.51 & 0.66$\pm$0.94 & 81.93\\
    & EOS & 0.60$\pm$0.85 & 0.36$\pm$0.58 & 0.48$\pm$0.55 &0.78 $\pm$0.93 & 86.14\\
    & Avg. & 1.05$\pm$1.19 & 0.15$\pm$0.43 & 0.60$\pm$ 0.64 & 1.16$\pm$1.33 & 81.52 \\
    \midrule
    \multirow{3}{*}{2.1 } & Ours & 0.52$\pm$0.84 & 0.48$\pm$0.84 & 0.60$\pm$0.63 & 1.01$\pm$0.98 & 82.53\\
    & EOS & 1.08$\pm$1.44 & 0.12$\pm$0.38 & 0.60$\pm$0.70 & 1.35$\pm$1.56 & 83.49 \\
    & Avg. & 1.45$\pm$0.95 & 0.48$\pm$0.84 & 0.96$\pm$0.71 & 1.62$\pm$0.93 & 82.77 \\
    \bottomrule
  \end{tabular}
  \label{ablation}

\end{table*}[htbp]

\subsection{Interpreting Bias Control and Generation Quality}
To qualify the effects of our method, we propose two different approaches to interpreting bias control and further evaluate the generation quality using standard metrics.

\textbf{Diffusion Attentive Attribution Maps} We visualize the cross-attention maps by profession of the generated images before and after debiasing. Note that we use the same prompts and generation settings across both conditions to ensure a fair comparison. Specifically, we adopt DAAM (Diffusion Attentive Attribution Maps)\cite{tang2022daam} to aggregate cross-attention maps across all denoising steps, allowing us to examine the cumulative influence of each token throughout the generation process. In particular, we focus on the attention distributions for the gender-neutral term \textit{person} and the profession phrases. While debiasing primarily targets gender attributes, it is important that profession-specific information is preserved. 

\begin{figure*}[!htbp]
     \centering
    \includegraphics[width=\linewidth]{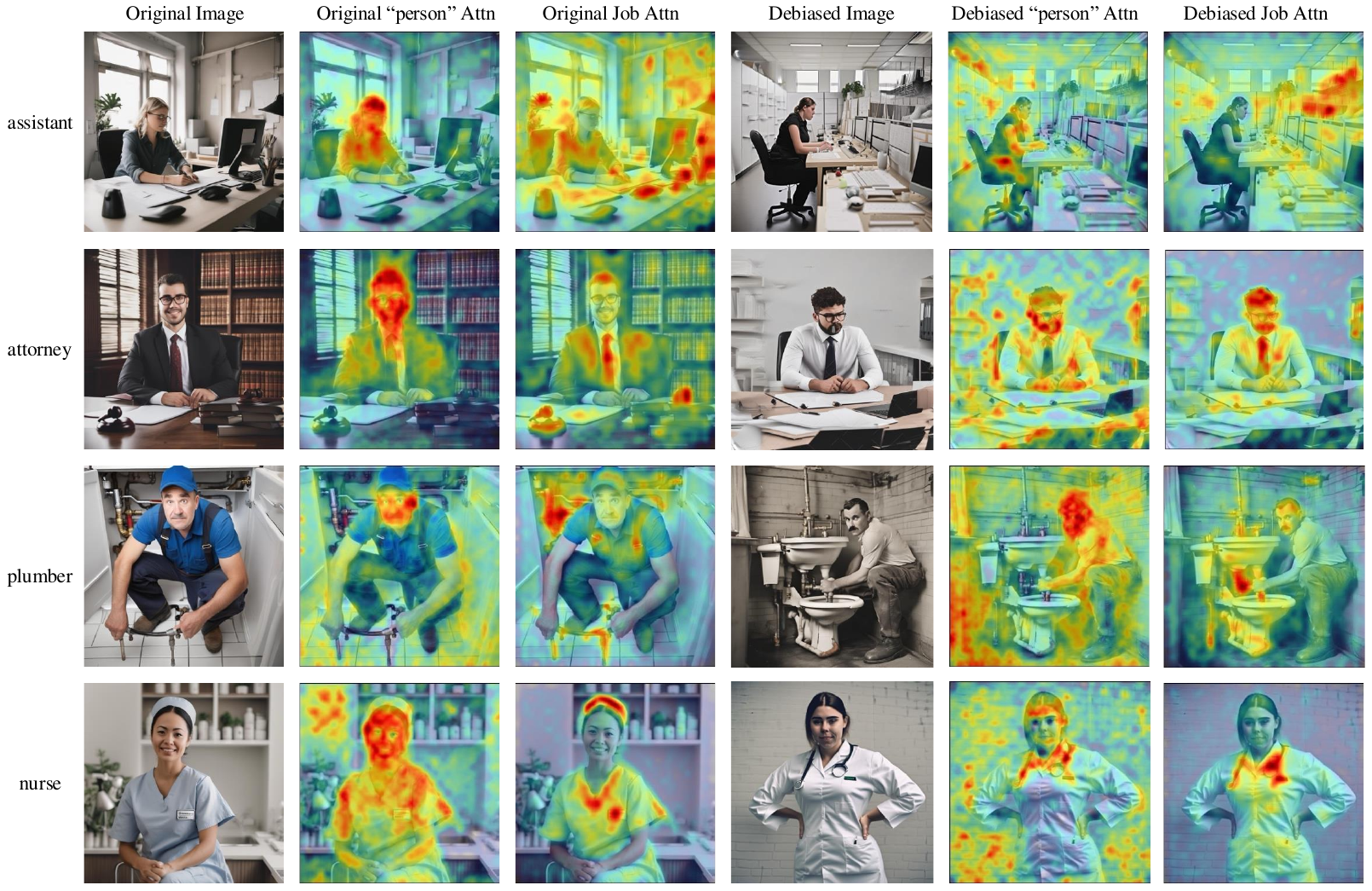}
    \vspace{-6mm}
  \caption{Visualization of attention maps before and after debiasing for different professions. Each row corresponds to a profession (Assistant, Attorney, Plumber, Nurse), showing the original photo, attention maps for the ``person'' token and the profession token, and their debiased counterparts.}
  \label{fig:attn_maps}
\end{figure*}

As shown in Figure ~\ref{fig:attn_maps}, we present several high-bias professions and analyze the attention associated with \textit{person} before and after debiasing. Prior to debiasing, the attention for \textit{person} is highly concentrated on the facial region, which is strongly correlated with gender attributes such as facial features and hairstyles. After debiasing, we observe that the attention becomes more diffused across the torso and the entire body and, in some cases, extends to elements relevant to the profession (e.g., tools, uniforms, or backgrounds). 
This attention dispersion indicates that the model relies less heavily on localized gender-specific features during generation. Importantly, the visual quality and profession-specific characteristics are maintained, suggesting that our method reduces gender bias without degrading semantic consistency or image fidelity. Additional qualitative examples are provided in the supplemental material to further support these findings.
\begin{table}
        \centering
        \scriptsize  % 更小的字体
        \setlength{\tabcolsep}{1.5pt}
    \caption{CLIP and IS scores before and after debiasing. Our method preserves semantic alignment and image quality(higher is better).}
      \vspace{1em}  % 你可以调成 0.5em、0.7em 试试效果
    \label{tab:clip_is}
    \begin{tabular}{lcccc}
    \toprule
    \textbf{Model} & \textbf{Original IS} & \textbf{Steered IS} & \textbf{Original CLIP} & \textbf{Steered CLIP} \\
    \midrule
    SD 1.4 & 16.10 & 15.72 & 21.13 & 21.13 \\
    SD 1.5 & 14.93 & 14.93 & 21.12 & 21.12 \\
    SD 2.1 & 14.69 & 14.70 & 21.12 & 21.12 \\
    SDXL  & 13.79 & 10.25 & 21.03 & 21.15 \\
    \bottomrule
    \end{tabular}
    \label{score}
\end{table}

\textbf{Text-Image Alignment and Visual Quality Evaluation} We further evaluate the generated images before and after applying SAE Debias across all model versions using two standard metrics: CLIP Score \cite{hessel2021clipscore} and Inception Score (IS) \cite{salimans2016improved}. As shown in Table \ref{score}, our method achieves comparable CLIP and IS scores before and after debiasing, suggesting that SAE Debias reduces gender bias without significantly affecting image quality or semantic alignment.
\section{Conclusion}
In this work, we propose a model-agnostic and lightweight framework for debiasing text-to-image diffusion models. We train a k-Sparse Autoencoder once on a large-scale gender-profession dataset to extract sparse, gender-relevant directions conditioned on profession. These directions are then used at inference time to steer and mitigate gender bias without retraining the base model. Our method, SAE Debias, outperforms both baseline Stable Diffusion models and existing debiasing techniques, achieving effective bias reduction while preserving semantic consistency and interpretability.

SAE Debias is plug-and-play, scalable across model versions and profession prompts, and offers a practical solution for mitigating occupational gender stereotypes in generative models. While our current work focuses on binary gender due to the structure of the Bias in Bios dataset, future extensions toward more inclusive gender representations are important directions for expanding fairness in generative AI.

{
    \small
    \bibliographystyle{IEEEtran}

    \bibliography{main}
}

\end{document}